# A lightweight target detection algorithm based on Mobilenet Convolution

Wang Shengquan

**Abstract:** Target detection algorithm based on deep learning needs high computer GPU configuration, even need to use high performance deep learning workstation, this not only makes the cost increase, also greatly limits the realizability of the ground, this paper introduces a kind of lightweight algorithm for target detection under the condition of the balance accuracy and computational efficiency, MobileNet as Backbone performs parameter The processing speed is 30fps on the RTX2060 card for images with the CNN separator layer. The processing speed is 30fps on the RTX2060 card for images with a resolution of 320×320.

## 1 Introduction

Target detection is important for understanding image content and finding target objects. This technique is essential in applications such as autonomous driving and augmented reality. In addition, real-time operation is also necessary, so a target detection model with fewer parameters is essential.

Since most devices have very limited computational power, it is important to build lightweight target detection models. Since Mobilenet convolution is small, the main approach to increase the perceptual field of the convolutional neural network is to replace Backbone with Mobilenet[1] using global pooling layers, and to reduce the size of the model by appropriate parameter reduction operations. The recognition speed of a single frame is improved.

## 2 Related studies

Since 1943, there has been a growing trend to use machines to achieve human-like perception, learning, memory and recognition. In 1985, Geoffrey-Hinton replaced the original single feature layer with multiple hidden layers and used the BP algorithm (back-propagation algorithm, proposed in 1969) to compute network parameters. In 1988, Kohonen et al. introduced the concept of neural networks. Subsequently, the basic theory of perceptrons and multilayer perceptrons was proposed, giving neural networks a certain basic model. Hubel and Wiesel found that the information processing process of biological vision is hierarchical through an in-depth study of the neural structure in biological neuroscience theory. By continuously extracting information between levels, the initial simple surface information is abstracted into higher-level feature information. Recognition of handwritten characters in postal codes using deep neural networks. Between 1987 and 1989, researchers proposed convolutional neural networks. Around 1995, researchers used physiology and computer technology to study vision problems. They proposed a sparse coding algorithm that iterated over 400 image segments to select the best segment weighting factor. However, the weights chosen were essentially the edges of images of different objects, which were similar in shape but different in orientation. In 2018, Redmon. J et al. proposed YOLOv3[5] , which is a little larger than YOLO9000[4] and YOLOv2 , but more accurate. Also the inference is faster. In 320x320 image detection, YOLOv3[5] has 28.2mAP in 22ms, which is as accurate as SSD, but three times faster.

## 3 Algorithms

### 3.1 Overall thinking

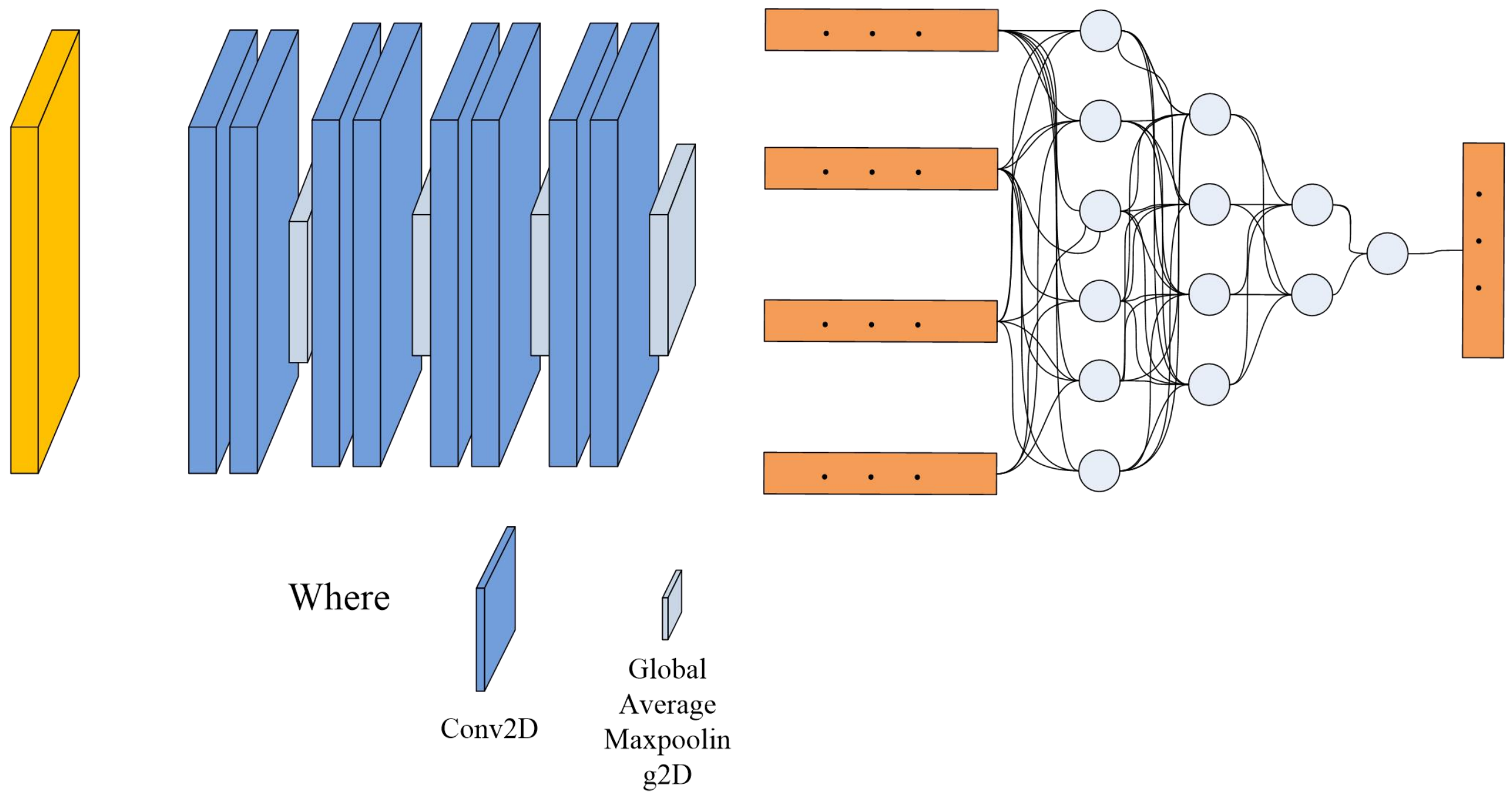

**Fig. 1 Convolutional neural network [1-6]**

The first step is to analyse the role of features in the process in conjunction with deep learning image processing algorithms, as feature extraction plays an important role in deep learning.

Suppose a matrix $I=\begin{bmatrix} I_1 & I_2 & I_3 & I_4 & I_5 \end{bmatrix}$ ,

where: $I_{\Theta j}=\begin{bmatrix} \beta_{11}+\alpha_{11}X & \ldots & \beta_{\Theta j}+\alpha_{\Theta j}X \end{bmatrix}$ ,

$\Theta \in \{0,1,3,\cdots\}, j=\{1,2,3\cdots\}$ $,I_j=\begin{bmatrix} I_{\Theta 1} \\ \vdots \\ I_{\Theta j} \end{bmatrix}$.

Since the convolution operation is more important in convolutional neural networks. Let the input matrix be the following:

$$I=\begin{pmatrix} I_{1,1} & I_{1,2} & I_{1,3} & I_{1,4} & I_{1,5} \\ I_{2,1} & I_{2,2} & I_{2,3} & I_{2,4} & I_{2,5} \\ I_{3,1} & I_{3,2} & I_{3,3} & I_{3,4} & I_{3,5} \\ I_{4,1} & I_{4,2} & I_{4,3} & I_{4,4} & I_{4,5} \\ I_{5,1} & I_{5,2} & I_{5,3} & I_{5,4} & I_{5,5} \end{pmatrix} \quad (1)$$

Set a convolution kernel of $3\times3$

$$k=\begin{pmatrix} kernal_{1,1} & kernal_{1,2} & kernal_{1,3} \\ kernal_{2,1} & kernal_{2,2} & kernal_{2,3} \\ kernal_{3,1} & kernal_{3,2} & kernal_{3,3} \end{pmatrix} \quad (2)$$

Convolution operations are performed with the input matrix using a convolution kernel, as follows:

$$c=conv2(I,k) \quad (3)$$

The pooling operation is then carried out:

$$p=globalaveragepool(c) \quad (4)$$

From Eqs. (1 - 3), which is a common procedure for convolutional neural networks, Eq. (4) indicates that the pooling operation is set to global mean pooling.

A typical convolutional neural network structure is shown in Fig. 1, but the main purpose of this structure is to extract the features better, and to increase the sampling rate while extracting the feature regions separately for subsequent global pooling.

The purpose of this is actually to distinguish between finer target features such as riders and pedestrians in road scenes where it is not enough to extract the appearance of the person, but the environment they are in is equally important. Experiments have shown that it is better to use classical convolutional neural networks.

However, downsampling has a great advantage. Filters operating on downsampled images have a larger receptive field, which allows them to collect more information. This is particularly important when differentiating between categories.

For example, in road scenes, it is not enough for the target networks of riders and pedestrians to understand the appearance of people, but their environment is equally important. Finally, it was found to be better to use Mobilenet convolution.

As mentioned above, it is important that the network has a wide receptive domain, so that it can perform classification by taking into account a richer set of relevant data and features. The study wanted to avoid excessive downsampling of the feature maps and decided to use Mobilenet convolutions to improve the model in the study. They replace the main convolutional layers in several bottleneck modules in a stage that operates at minimal resolution. These greatly improved the accuracy of the operations, by about 4 percentage points, with no additional loss.

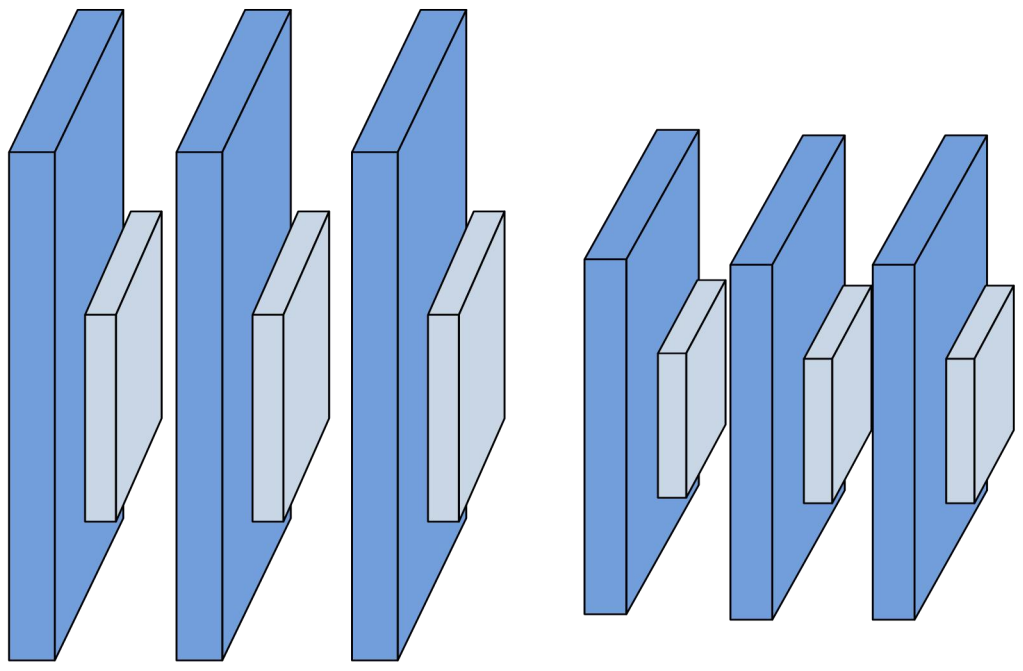

**Fig. 2 Simplified Darknet diagram**

As shown in Figure 2, the structure of Darknet53 is a combination of DarknetConv2D, a deep network with 147 layers, which is effective in detecting small targets with high accuracy and speed, and is used in well-known target detection algorithms such as YOLOv3.

| Layer | Filter size | Repeat | Output size |
|---|---|---|---|
| Image | | | 416*416 |
| Conv | 32 3*3/1 | 1 | 416*416 |
| Conv | 64 3*3/2 | 1 | 208*208 |
| Conv | 32 1*1/2 | | 208*208 |
| Conv | 64 3*3/1 | 1 | 208*208 |
| Residual | | | 208*208 |
| Conv | 128 3*3/2 | 1 | 104*104 |
| Conv | 64 1*1/1 | | 104*104 |
| Conv | 128 3*3/1 | 1 | 104*104 |
| Residual | | | 104*104 |
| Conv | 256 1*1/2 | 1 | 52*52 |
| Conv | 128 1*1/1 | | 52*52 |
| Conv | 256 3*3/1 | 1 | 52*52 |
| Residual | | | 52*52 |
| Conv | 512 3*3/2 | 1 | 26*26 |
| Conv | 256 1*1/1 | | 26*26 |
| Conv | 512 3*3/1 | 1 | 26*26 |
| Residual | | | 26*26 |
| Conv | 1024 3*3/2 | 1 | 13*13 |
| Conv | 512 1*1/1 | | 13*13 |
| Conv | 1024 3*3/1 | 1 | 13*13 |
| Residual | | | 13*13 |

**Fig. 3 Darknet structure diagram**

Each network was trained with the same settings, and all were tested with images from $256 \times 256$ , and all were single precision. Figure 4 shows the interpretation of DarknetConvolutional2D_BN_Leaky in Figure 2, and Figure 5 shows the performance of some of the current mainstream target detection algorithms on the open source dataset.

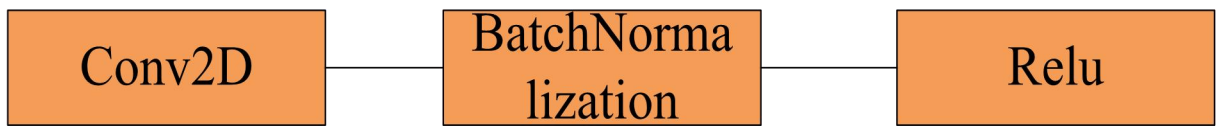

**Fig. 4 DarknetConvolutional2D _ B N _ Leaky**

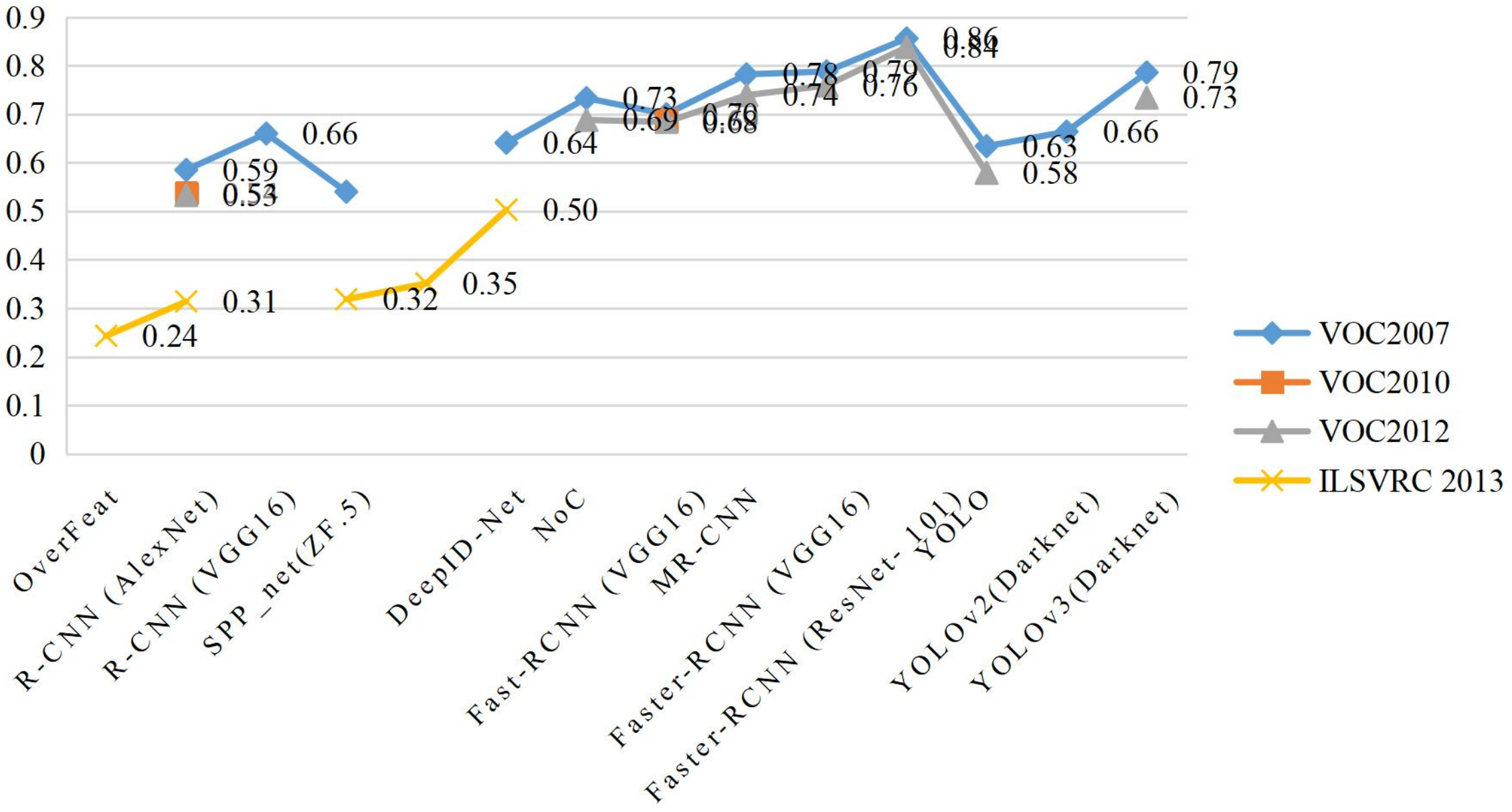

**Fig. 5 The performance of the current mainstream target detection algorithms on VOC and ILSVRC, where the ordinate refers to the accuracy**

The model mentioned in this paper is improved by replacing the backbone with Mobilenet, which reduces the overall computational effort of the network.

This is done by balancing accuracy and computational efficiency. In terms of targeted training and optimization, the following work has been done in this paper:

MobileNet as Backbone, and parameter optimisation of Backbone

Increase sampling on Mobilenet to improve the recognition rate of small features

Increasing the perceptual wildness of the CNN partial separator layer by using a global pooling layer

Enhanced for small sample data.

As a result of these operations, the structure in Figure 6 is obtained.

### 3.2 Relevant explanations

#### 3.2.1 Using global mean pooling to improve upsampling

There is a project for YOLOv3 on GITHUB, which uses a lot of downsampling for the tuning. Since it is only tested on the PASCAL VOC dataset, which has large targets with distinctive features, the overall improvement is made by adding an upsampling layer using a global pooling layer, which replaces the global average pooling layer with a global average pooling layer, while keeping the basic accuracy unchanged. This is to enable the classification methods with different labels to generate different feature maps.

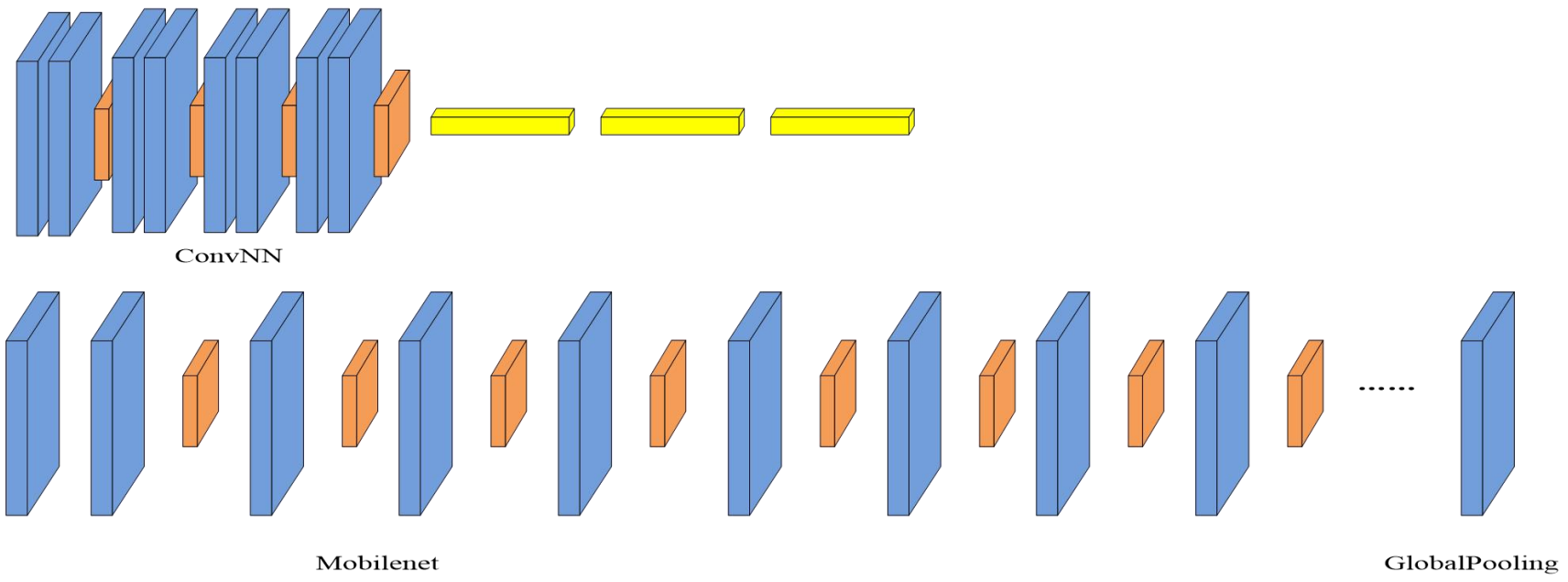

**Fig. 6 Convolutional neural network + Mobilenet**

### 3.2.2 Mobilenet

Mobilenet is an extremely simple network structure, with small network depth and fast inference. It is also the first of its kind for migration networks. MobileNet is a depthwise separable convolution [7] model, which uses depthwise separable convolution to convolve its sub-channels and uses only the convolution kernel of 1×1 for each convolution, which can greatly reduce the number of parameters in the model. The structure of Mobilenet is shown in Figure 7.

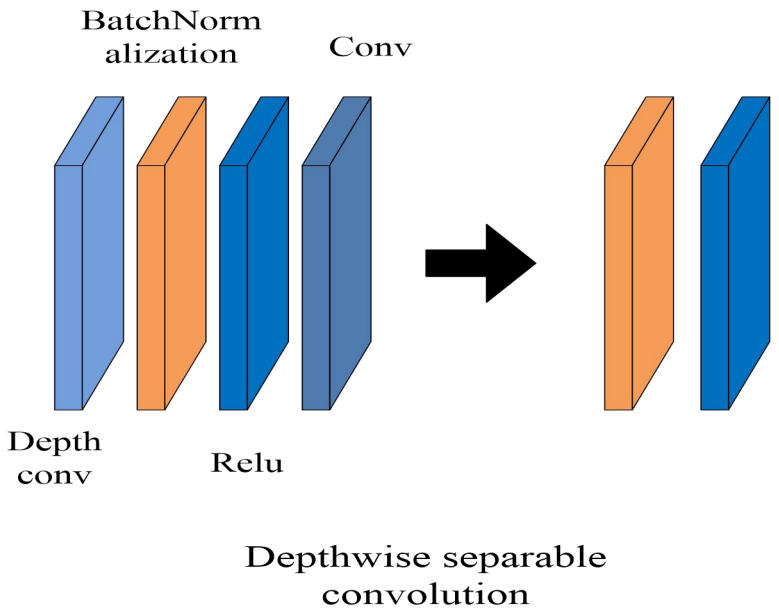

**Fig. 7 The Mobilenet Simplified Structure**

### 3.2.3 Reduction of parameters

The number of parameters determines to a large extent the size and learning ability of the model, but of course it is not a matter of increasing, let alone decreasing, the number of parameters. This limit should be determined by balancing accuracy and model size. In this paper, the parameters are reduced by 1/4 of the size, but the depth of the algorithm does not change as a result. In fact, since the algorithm was originally designed to understand typical targets such as pedestrian recognition, they do not necessarily need a large number of parameters to improve the learning ability, but only a certain number of parameters to extract the features, which is not only better for the task of target detection. This is not only better for target detection, but also for reducing the size of the model.

## 4 Training

Due to the significant disproportion between small and large samples in the dataset, the data were normalized by removing the very sparse labels and augmented with a generative adversarial network[8] (GAN) and denoised with a wavelet algorithm. The cost function used is Eq. (5), where S is the number of all grids, B is the number of predicted borders, and x, y, w, h, c are the five calibration values.

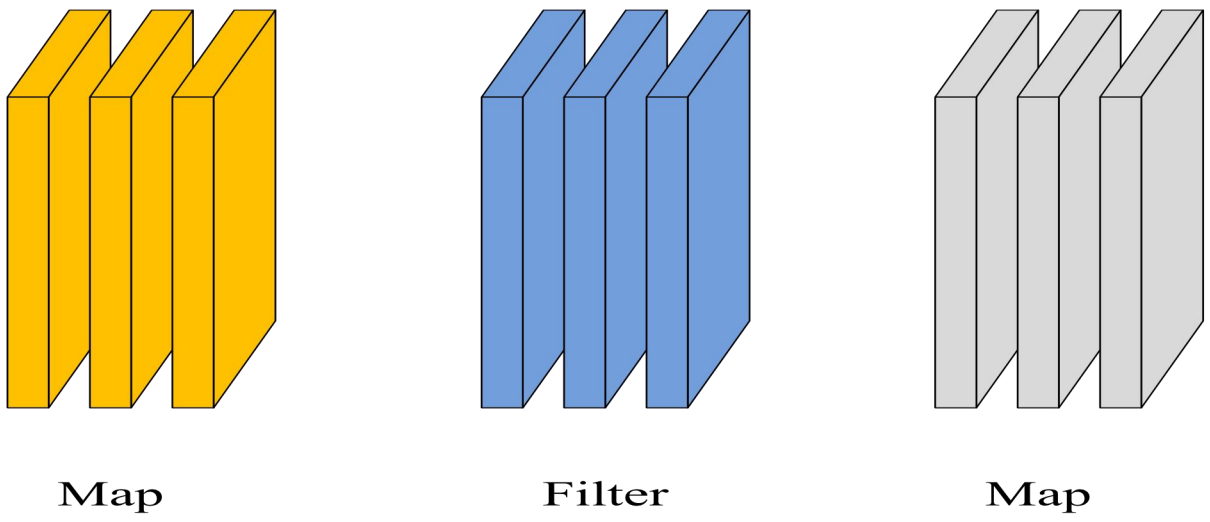

**Fig. 8 Depthwise separable convolution**

$$
\begin{aligned}
L(x,y,w,h,c) = {} & \lambda_{\text{coord}} \sum_{i=0}^{s^2}\sum_{j=0}^{B} l_{i,j}^{ob}\left[(x_i-\bar{x}_i)^2+(y_i-\bar{y}_i)^2\right] \\
& +\lambda_{\text{toord}} \sum_{i=0}^{s^2}\sum_{j=0}^{B} l_{l,j}^{obj}\left[\left(\sqrt{w_i}-\sqrt{\hat{w}_i}\right)^2+\left(\sqrt{h_l}-\sqrt{\hat{h}_i}\right)^2\right] \\
& +\sum_{i=0}^{s^2}\sum_{j=0}^{B}\left[l_{i,j}^{bb}\left(C_i-\bar{C}_i\right)^2+\lambda_{\text{hoob}} l_{i,j}^{\text{noobj}}\left(C_i-\bar{C}_i\right)^2\right] \\
& +\sum_{i=0}^{s^2} l_i^{0bj}\sum_{c\in\text{classes}}\left(p_i(c)-\hat{p}_i(c)\right)^2
\end{aligned} \tag{5}
$$

The overall structure is shown in Figure 9: Since the computer configuration used for the experiments was far from the level of a standard deep learning lab's arithmetic workstation, the size of the input images was reduced from the original design to a uniform size of . The training data was obtained from PASCAL VOC images on a computer with an I5 processor, 16 GB of RAM, and an RTX2080 graphics card, while the weights from the training tests were obtained using an RTX2060 graphics card. Tests were also conducted on a GTX1050TI laptop, but as there is still a gap between the graphics cards of laptops and desktops, it is not recommended to use a laptop platform. It is not recommended to use a laptop platform for testing.

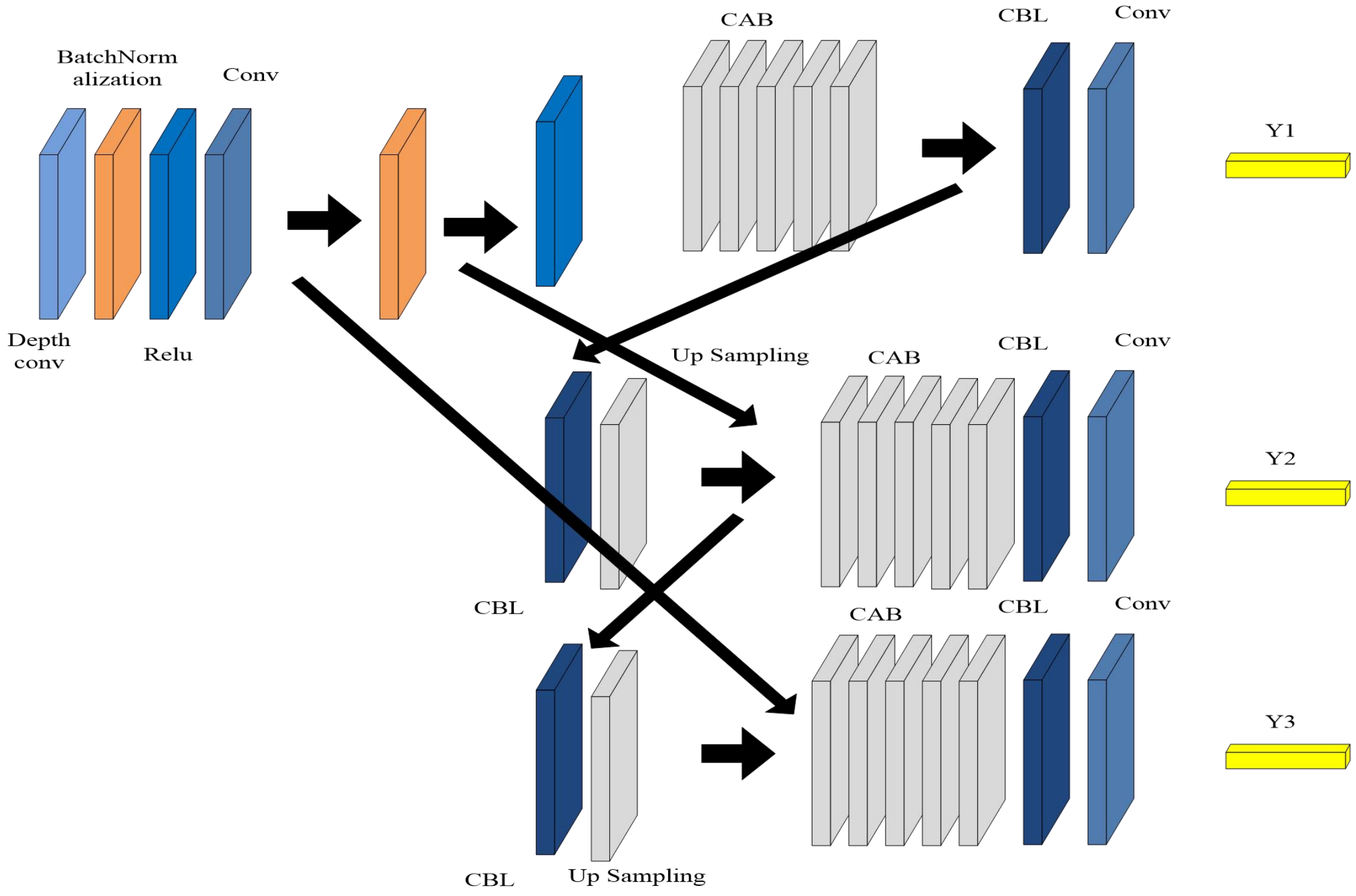

**Fig. 9 Overall Structure**

As shown in Figure 10, Mobilenet is used as the Backbone of the grafting network, which contains a combination of modules as shown in Figure 4, including a number of positive feedback connections, the data set is fed into the network and trained 1000 times, including Training, Trainingval, Test and other processes, and finally the corresponding weight file is obtained The test is ready for testing.

As can be seen, the structure has an upsampling component, which is actually necessary for the current mainstream YOLO target detection model, to improve the quality of the detection by increasing the upsampling to obtain the desired target frame. Although the experimental metrics produced by the method are comparable to YOLOv3, and even slightly behind in AP50 and AP, the weight size is significantly reduced, due to the above-mentioned reduction in the number of parameters and the use of a mean pooling layer in Mobilenet with increased upsampling to reduce the model size.

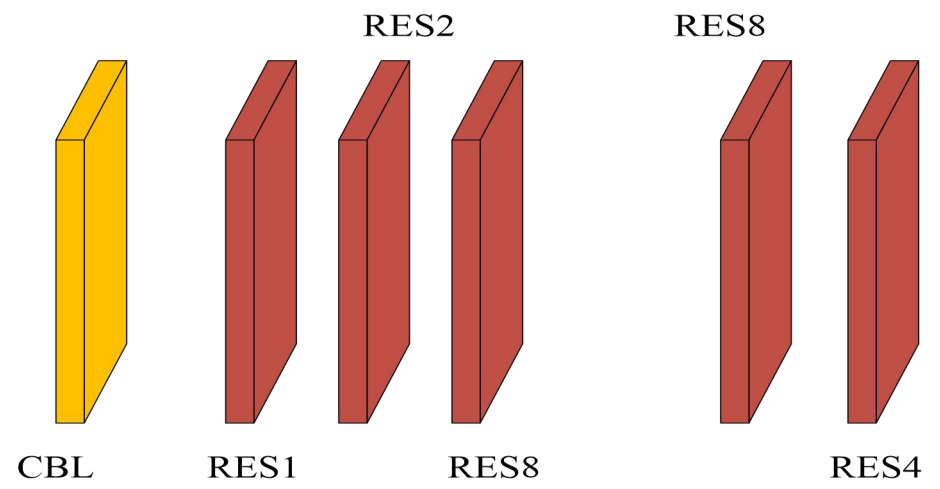

**Fig. 10 The Darknet53 replaced by Mobilenet**

Figure 10 shows a simplified structure of part of Figure 3, mainly to explain the Darknet 53. that Mobilenet has replaced.

Res is the residual block, CBL is introduced in Figure 4.

Darknet53 has a very large number of participants, while Mobilenet is very small.

In addition, the smaller GPU memory can also be used to set a larger Batchsize during training, thus improving the training efficiency.

Since Batchsize determines the number of input images in a single training batch, it would make sense to increase its size with limited memory.

The performance of this target detection model is not very different from the rest of the models. On the contrary, its size is smaller than most models. And its inference speed is 30 FPS.

The results for the smaller samples are worse than the others, but the rest of the results are better and do not differ much, which is reasonable due to the small number of parameters and the simple structure, and will be improved if the number of training sessions increases.[9-16] .

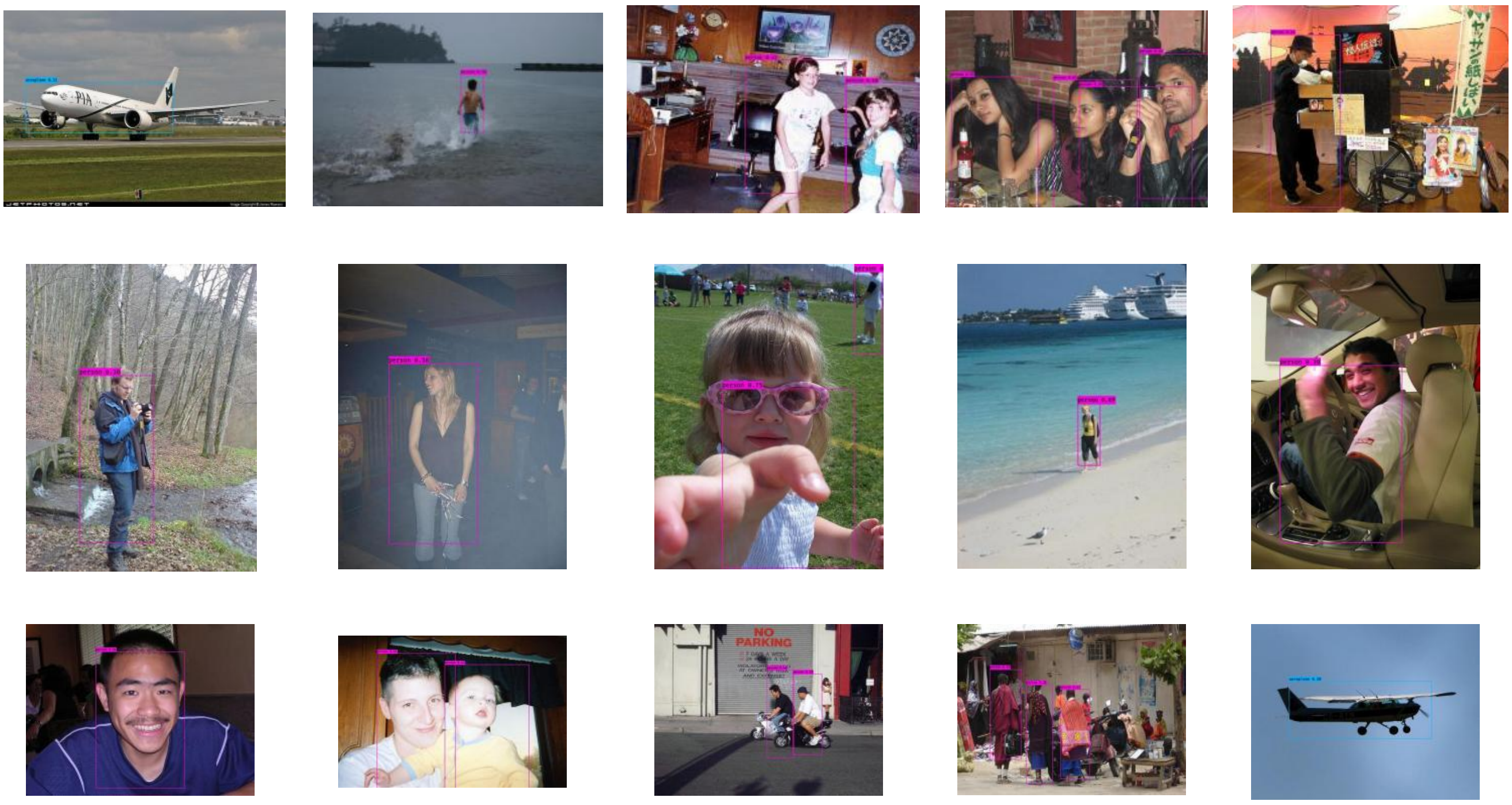

**Fig. 11 In the PASCAL VOC test set test, in the PASCAL VOC test effect, it can be very intuitive to see, in the large sample identification effect is very good**

Figure 12 shows a simplified diagram of the structure of a discrete graphics card, with hundreds of stream processors in green and the memory module in yellow. The memory usage during training is in accordance with the following equation.

$$RAM_for_training = Modelsize + batchsize \times N \tag{6}$$

Obviously, the 28.7mb weight file reference is smaller than 230mb and 150mb, so the model has a significantly lower footprint in the video memory and thus has some advantages for training devices with low video memory.

## 5 Conclusion

The efficiency of feature extraction has been improved and the chance of overfitting has been reduced by using parameter reduction. With these improvements, the network is able to improve the speed of target extraction and target recognition in images. Of course, the accuracy of this algorithm is somewhat affected by the fact that it is not as deep as traditional deep learning algorithms, with a size of only 28MB, but this can be solved by increasing the number of iterations and the dataset. When training the classical dataset, we did not have time to investigate how the classical network was augmented due to the time and equipment impact. The classical dataset VOC PASCAL has a large number of small samples of images, so some augmentation is necessary to improve the results.[17] The test case achieved good results as the number of labels were all collected in a near 1:1 ratio. Finally, although the target detection algorithm described in this paper is not very advantageous to a certain extent, the idea of migrating the training and solving the overfitting problem by reducing the number of parameters is of great importance in future research, and many algorithms are currently validated by such methods.

Table　1 Relevant indicators

**Tab. 1 Related indicators**

| | AP / % | AP50 / % | AP75 / % | APs / % | APM / % | APL / % | FPS / % | Size / *mb* |
|---|---|---|---|---|---|---|---|---|
| Two-stage model | | | | | | | | |
| Faster R-CNN [2] | 34.9 | 55.7 | 37.4 | 15.6 | 38.7 | 50.9 | 5-17fps | ≈150*mb* |
| Faster R-CNN+ FPN [2] | 36.2 | 59.1 | 39.0 | 18.2 | 39.0 | 48.2 | | |
| Faster R-CNN +G-RMI [2] | 34.7 | 55.5 | 36.7 | 13.5 | 38.1 | 52.0 | | |
| Faster R-CNN +TDM [2] | 36.8 | 57.7 | 39.2 | 16.2 | 39.8 | 52.1 | | |
| Single stage model | | | | | | | | |
| YOLOv3 | 21.6 | 44.0 | 19.2 | 5.0 | 22.4 | 35.5 | 28-30fps | ≈ 230*mb* |
| Ours | 20.5 | 40.3 | 30.6 | 6.9 | 24.6 | 35.9 | 30fps | ≈ 28.7*mb* |

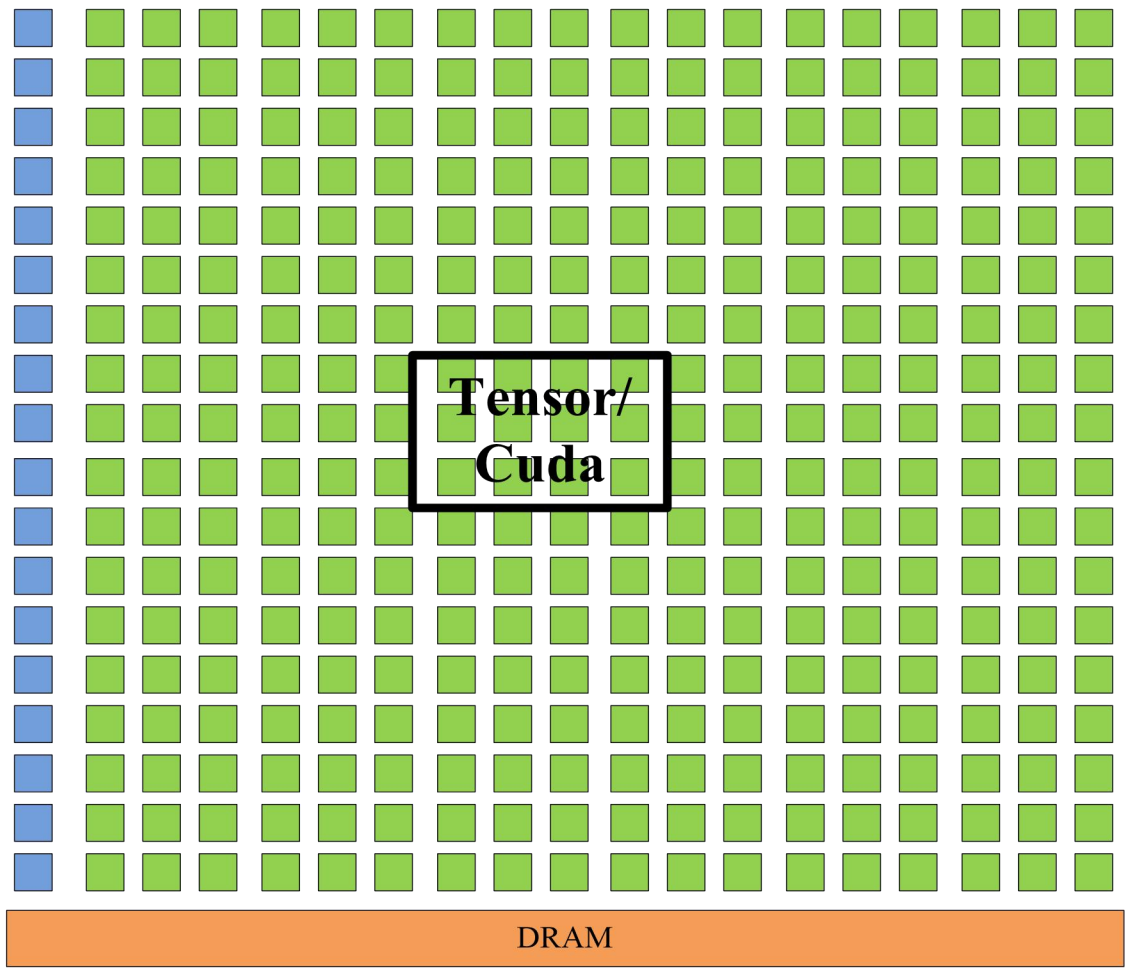

**Fig. 12 Structural simplified diagram of the independent**